# Motion Planning with Gamma-Harmonic Potential Fields

Ahmad A. Masoud
Electrical Engineering Department, King Fahad University of Petroleum and Minerals, Dhahran 31261, Saudi Arabia,
e-mail: masoud@kfupm.edu.sa

*Abstract:* This paper extends the capabilities of the harmonic potential field (HPF) approach to planning. The extension covers the situation where the workspace of a robot cannot be segmented into geometrical subregions where each region has an attribute of its own. The suggested approach uses a task-centered, probabilistic descriptor of the workspace as an input to the planner. This descriptor is processed, along with a goal point, to yield the navigation policy needed to steer the agent from any point in its workspace to the target. The approach is easily adaptable to planning in a cluttered environment containing a vector drift field. The extension of the HPF approach is based on the physical analogy with an electric current flowing in a nonhomogeneous conducting medium. The resulting potential field is known as the gamma-harmonic potential (GHPF). Proofs of the ability of the modified approach to avoid zero-probability (definite threat) regions and to converge to the goal are provided. The capabilities of the planner are demonstrated using simulation.

Keywords: probabilistic robotics, autonomous agents, motion planning, navigation, risk minimization, harmonic potential

Nomenclature

| | |
|---|---|
| HPF: | Harmonic potential field |
| GHPF: | Gamma-harmonic potential field |
| BVP: | Boundary value problem |
| LFC: | Liapunov function candidate |
| P(x): | a probabilistic field describing of the fitness of the point x to carry-out a task |
| $\Omega$: | workspace of the agent |
| $\Omega_o$: | definite threat (P(x)=0) subset of $\Omega$ |
| $\Omega_a$: | the admissible region of the probabilistic workspace ($\Omega_a = \Omega - \Omega_o$) |
| O: | forbidden regions in deterministic worspaces |
| $\omega$: | an infinitesimally small set in $\Omega$ |
| u(x): | navigation policy |
| $x_s$: | starting point of motion |
| $x_T$: | target point |
| $\phi$: | the empty set |
| $\rho(t)$: | the time paramerized trajectory laid by the planner |
| H(x): | the hessian matrix |
| $\nabla$: | the gradient operator |
| $\nabla^2$: | the laplacian operator |
| $\nabla \cdot$: | the divergence operator |
| $\Gamma$: | boundary of the deterministic workspace |
| V(x): | potential field |
| $\sigma(x)$: | conductivity |
| J(x): | electric current density |
| $\Psi(x)$: | a vector field describing the drift force in the workspace |
| $\Xi$: | Liapunov function |
| $\dot{\Xi}$: | time derivative of $\Xi$ |
| E: | minimum invariant set |

# I. Introduction:

Designing an autonomous agent is a challenging multi-disciplinary task [1]. Special attention is being paid to the propulsion, data acquisition and communication systems used by an agent. However, the biggest challenge seems to be in designing a proper planning module. The function of this module is to unite these sub-systems into one goal-oriented unit. There is a long list of conditions a planner must satisfy. These conditions are necessary in order to generate a sequence of instructions which the actuators of motion may execute to successful completion of an assignment. However, the conditions on handling and representing mission data seem to be the most stringent [2]. There are two core requirements a representation should satisfy. They are

1- compatibility with the manner in which data is being processed and action is being generated,

2- updatability of the representation.

Updatability requires that the validity of the existing portion of the representation not be conditioned on the future data that could be received. This allows for the incremental construction of a representation. Another important issue is to increase the diversity of environment-related, operator-supplied information which the planner is capable of processing in order to yield the guidance signal.

Most planners assume divisible environments that may be partitioned into subsets of homogeneous attributes. The most common scheme is to have a binary partition of admissible sets and forbidden ones. Binary partitions may be constructed using geometric structures like circles [3], occupancy maps, Voronoi partitions [4], grids and graphs [5], samples [9] and trees [6]. There are situations where the environment in which an agent is operating is not divisible. For example, a plane flying through turbulent atmosphere [7] experiences a degree of turbulence where clear space is diffused into turbulent space with no sharp boundaries separating the two. Also, in the case of mobile robots operating in rough terrains [8], the description should be based on the degree of difficulty in negotiating the terrain. It is highly unlikely that success can be attained by basing the actions on a binary geometric partition of the environment into admissible and forbidden regions.

An alternative to the geometric approach is the use of a soft representation that consists of a field reflecting, at each point in the environment, the probability of achieving the task. Probabilistic representations are ideal for encoding the information in non-divisible environments. These representations may also be used to efficiently incorporate spatial ambiguity and the aging of information in the representation. For example, exact knowledge of the location of a one-dimensional point object may be represented as a probability distribution function (PDF) with an impulse function (figure-1). Convolution of this exact spatial knowledge with the proper blurring operator produces another representation with the ambiguity factored-in. Repeated application of the blurring operator leads to a uniform PDF representing the maximum state of ambiguity (i.e. the object could be anywhere).

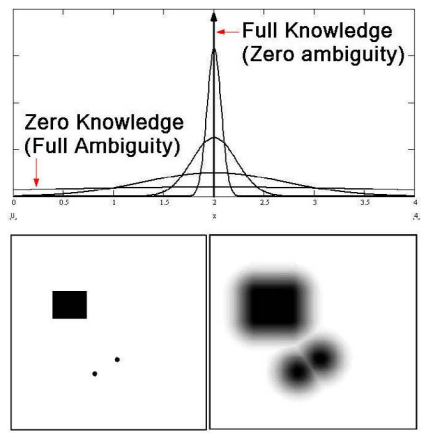

Figure-1: spatial ambiguity and aging of information

Errors and uncertainty in the sensed signals are almost always present in navigation systems. Countermeasures [38] to reduce their effect on the generated trajectory have to be incorporated into the system for robust operation. Soft probabilistic representations are an effective approach to counteracting ambiguity. They have been adopted by many researchers in robotics. For example, in [10,11], probabilistic representations were used for path planning in the presence of sensory data ambiguity. This representation may be fed to a reinforcement learning-based, or an optimal control-based stage, to generate the navigation policy. It is noticed that the structure of the ambiguity field has a pronounced influence on the performance of the policy generator. The less structured the field is, the more time it takes the algorithm to converge. In [39], the ambiguity in determining the location of a target moving in an obstacle-free space was probabilistically modeled using an occupancy map. This map is fed into the potential field stage to direct the motion of the pursuer in a direction that would increase the probability of capturing the target. Fuzzy logic techniques [12] are used to derive navigation actions for robots based on soft probabilistic representations. While these techniques have proven their practicality, they are not provably correct (heuristics). Qualitative methods for dealing with uncertainty may also be found in [13].

Occasionally, a workspace contains a force external to the agent that influences its state. This force is known as the drift field. A drift field is usually treated as a source of disturbance whose influence should be suppressed by the agent's low-level controller. With the advances in forecast technology [14], a drift field may be predicted for a considerable period ahead of time. These advances make it possible to consider drift as a source of information which the planner should take into consideration instead of being a source of disturbance that has to be suppressed.

Being able to treat drift as a source of information has important applications in the area of energy-exhaustive missions. When energy is a scarce resource, good planning does not only reduce the drain caused by drift, but may even use it as a source for powering the agent. In the past, several techniques were suggested for incorporating drift fields in the planning process. Most of these techniques use optimization or search-based methods for determining the minimum cost path that connects two points in the drift-occupied space while avoiding the obstacles. In [15], a genetic algorithm planner is used to lay a minimum energy trajectory for an AUV operating in turbulent waters. A tree-based trajectory planner [16] is used to plan a path for an aerial glider so that the path is laid along the component of the wind which exerts maximum lift. The A* search approach is used for planning a path for an AUV operating in a current field [17]. A symbolic wave expansion approach is developed to tackle planning in workspaces with dynamic current fields [18]. Other approaches may be found in [19,20].

This study suggests a method for generating the navigation policy for a robot using a goal point, and a task-centered, probabilistic description of the environment, as inputs. The method is efficient, provably-correct and assumes no structure whatsoever on the descriptor used to represent the environment. The developed method can be adapted in a straightforward manner to motion planning in a cluttered workspace that is populated by a drift field.

The suggested approach is called the gamma-harmonic [33] potential field planning approach. It is derived by extending the capabilities of the deterministic harmonic potential field (HPF) approach to motion planing [21,22,23] in order to tackle environments that can only be probabilistically described. Previous attempts to construct a probabilistic HPF planner focus on using HPF as a probability measure describing the danger of collision with obstacles in the environment. This strips the HPF approach from its well-justified role as a navigation policy generator and reduces it to a merely descriptive tool that is an input to a navigation policy generation stage [24,25].

This paper is organized in the following manner. In section II, the problem is stated along with its scope. In section III, the probabilistic, HPF-based planning approach is developed. In section IV, an interpretation of the task-centered, probabilistic descriptor is suggested for incorporating drift in the trajectory generation process. Section V provides a basic analysis of the behavior of the approach. Simulation results are in section VI and conclusions are placed in section VII.

## II. Problem Statement and Scope

The objective of this work is to develop a navigation guidance policy (u) generator that can be used by a point mobile agent to convert an *a priori* known, differentiable probabilistic field (P) that describes the content of the workspace ($\Omega$) and a target point ($X_T$) into a well-behaved vector field that may be used by the agent to direct its motion in $\Omega$. The probabilistic field describes, in a task-centered manner, the fitness of a point x in $\Omega$ to carry-out the intended task. Let $\Omega o$ ( $\Omega o \subset \Omega$) be the set of definite threat region defined

as
$$\Omega o = \{x : P(x) = 0\}. \tag{1}$$
regardless of the structure of P, the first order dynamical system that is constructed using the navigation policy
$$\dot{x} = u(x, P(x), x_T) \qquad \forall \, x(0) \in \Omega - \Omega o \tag{2}$$
must yield a goal-seeking guidance action that enables a point agent to converge from anywhere in Ω-Ωo to the target
$$\lim_{t \to \infty} x(t) \to x_T. \tag{3}$$
The navigation policy must also guarantee that definite threat regions (Ωo) are avoided at all times:
$$x(t) \cap \Omega o \equiv \varphi \qquad \forall t. \tag{4}$$
The trajectory ρ(t) which is the solution of the dynamical system in (2) is required to be analytic. It is also required that the risk (risk=1/P(x)) accumulated by moving along this trajectory
$$\int_0^\infty \frac{1}{(P(\rho(t)))} |\dot{\rho}(t)| dt \tag{5}$$
be reduced to a global minimum.

To the best of this author's knowledge, the suggested navigation policy generator is the first to directly generate a goal-seeking, guidance action from a probabilistic description of the environment. The approach amasses a considerable number of novel and important properties:

1- it can process inherently ambiguous data while making provably-correct statements on the resulting behavior of the agent,

2- the probabilistic descriptor has an intuitive format whose local effect on the generated action is discernable by the operator. This makes the policy generator a valuable component in a decision support system [40],

3- the probabilistic descriptor need not be generated from an underlying deterministic model of the environment. Therefore, its use extends beyond that of modeling the ambiguity and uncertainty of a deterministic model to dealing with inherently stochastic environments such as rough terrains,

4- the goal-seeking guidance action adds to the robustness of the planning process in the sense that if a disturbance throws the agent off-course, the new location contains another sequence of instructions that will lead the agent to the target. The analyticity of the solutions making up the navigation policy causes the deviation from the original path to be commensurate with the amount of disturbance the agent is subjected to,

5- forcing the solution trajectories to be analytic results in well-behaved, dynamically-friendly paths that can, with high probability, be converted by the control unit of the agent into a navigation control,

6- as will be demonstrated in this paper, the nature of the probabilistic descriptor, and the fact that it doesn't have to satisfy any condition other than differentiability, have significant advantages. They make it possible, with little effort, to change the format of the environment data to suit other challenging planning problems, e.g., planning in the presence of drift fields for energy exhaustive missions.

## III. The Gamma-Harmonic Approach

**III.1: HPF: A Background**

Harmonic potential fields (HPFs) have proven to be effective tools for inducing, in an agent, an intelligent, emergent, embodied, context-sensitive and goal-oriented behavior (i.e. a planning action). A planning action generated by an HPF-based planner can operate in an informationally-open and organizationally-closed mode [26]. Thus, an agent is able to make decisions on-the-fly using on-line sensory data without relying on the help of an external agent. HPF-based planners can also operate in an informationally-closed, organizationally-open mode. This mode makes it possible to utilize existing data about the environment in generating the planning action as well as eliciting the help of external agents in managing the task at hand. Such features make it possible to adapt HPFs for planning in a variety of situations. For example, in [27] vector-harmonic potential fields were used for planning with robots having second order dynamics. In [28], the approach was configured to work in a pursuit-evasion planning mode, and in [29], the HPF approach was modified to incorporate joint constraints on regional avoidance and direction. The decentralized, multi-agent planning case was tackled using the HPF approach in [30]. The HPF approach was also found to be able to facilitate the integration of planners as subsystems in networked controllers [31]. It effectively addresses the real-life requirements needed to successfully network sensory, communication and control modules. A basic setting of the HPF approach is shown in (6).

Solve:
$$\nabla^2 V(x) \equiv 0 \quad X \in \Omega \tag{6}$$
subject to: $V(x_S) = 1$, $V(x_T) = 0$, and $\dfrac{\partial V}{\partial \mathbf{n}} = 0$ at $x = \Gamma$,

A provably-correct path may be generated using the gradient dynamical system

$$\dot{x} = -\nabla V(x). \tag{7}$$

where $\Omega$ is the workspace, $\Gamma$ is its boundary, $\mathbf{n}$ is a unit vector normal to $\Gamma$, $x_s$ is the starting point, and $x_T$ is the target point [29].

**III.2: A Physical Metaphor:**

Equations (6) & (7) model the behavior of an electric current flowing in a homogeneous conductor with a conductivity $\sigma(x)$= constant [32]. The conductor is populated by perfect insulators ($\sigma=0$) occupying the forbidden regions surrounded by $\Gamma$ (figure-2).

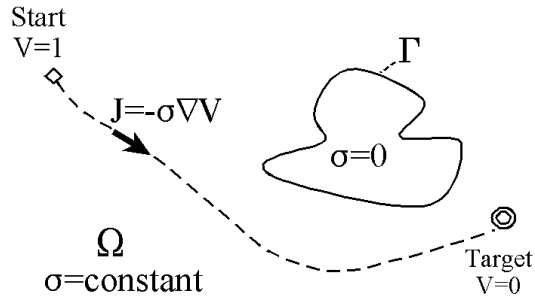

Figure-2: Physical metaphor for the planner

The flow is described by the electric current density(J)

$$J(x) = -\sigma(x)\nabla V(x). \tag{8}$$

The homogeneous Neumann condition means that the current cannot penetrate the perfect conductor and has to move tangentially to it. The Laplce equation is simply a product of applying the continuity condition to the electric flow
$$\nabla \cdot (-\sigma(x)\nabla V(x)) = 0. \tag{9}$$

If $\sigma$ is constant, equation (9) reduces to the well-known Laplace operator in (6). The conductivity $\sigma$ may be used to represent how favorable a point in the robot's space is to conducting motion. A $\sigma=0$ means that the corresponding space does not support motion at all. On the other hand, a high value of σ means that the corresponding space highly favors motion.

**III.3 The Gamma-Harmonic Planner**

It is possible to establish a tight analogy between the above situation and the situation where the agent's environment is to be represented in a task-centered manner using a descriptor (P(x)). This descriptor specifies at each point in the space the agent's ability to perform the assigned task. It does not matter what the causes are e.g. sensor problems, rough terrains, man-made hazzards etc. If the agent is expected to encounter difficulties operating in a certain region of space, a low value for P(x) is assigned to that region. A navigation policy that satisfies the conditions stated in II may be generated as

$$\nabla \cdot (P(x)\nabla V(x)) \equiv 0 \qquad x \in \Omega \qquad (10)$$

subject to $\qquad V(x_S) = 1, \ V(x_T) = 0$

A provably-correct, analytical trajectory may be constructed using the gradient dynamical system

$$u(x) = -\nabla V(x). \qquad (11)$$

The potential field (V) generated by (10) is known as the gamma-harmonic potential field (GHPF) [33].

The GHPF operator in (10) is strongly related to the ordinary Laplace operator and possesses a useful physical interpretation. First, the operator should be expanded as

$$\nabla \cdot (P(x)\nabla V(x)) = P(x)\nabla^2 V(x) + \nabla P(x)^t \nabla V(x)) = 0, \qquad (12)$$

which leads to $\qquad \nabla^2 V(x) = -(1/P(x))(-\nabla P(x)^t(-\nabla V(x))).$

Notice that $-\nabla V(x)$ is the direction at which motion is to be driven and $-\nabla P(x)$ is a vector pointing in the direction of increasing risk. Also, keep in mind that the Laplacian of a potential is the divergence of the gradient field which is physically defined as the outflow of flux when the volume shrinks to zero. If the Laplacian is negative, there is a sink in the closed region, i.e. motion is indebted by absorbing the gradient flux used to direct motion. If the Laplacian is positive, there is a source in the closed region, that is, motion is stimulated by aiding the flow of the gradient flux. When the Laplacian is zero (Lapace equation), the region is neutral towards the gradient flux. As can be seen, the GHPF operator may be viewed as an intelligent version of the HPF that is sensitive to the future ability of the agent to carry-out its task (figure-3).

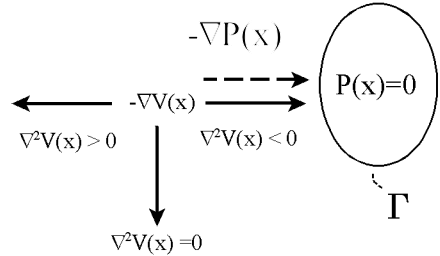

Figure-3: The modified operator exhibits intelligent behavior.

## IV. Adapting the planner to the drift case

It is shown that $P(x)$ may be easily interpreted in an opportunistic manner that exploits drift as a source for powering the agent.

A planner that mitigates the effect of drift on motion is expected to satisfy the following conditions

$$\lim_{t \to \infty} x(t) \to x_T, \qquad \forall x(0) \in \Omega$$
$$x(t) \cap O \equiv \phi \qquad \forall t$$
$$U = \int_0^\infty Fc(\Psi(x(t)), -\nabla V(x(t))) |\nabla V(x(t))| dt$$

(13)

where the functional U has to be minimized, or reduced, to a satisfactory value. $x \in R^N$, O is the set of forbidden regions (obstacles, $\Gamma = \partial O$), $\Omega$ is the subset of admissible space (worksapce), $\Psi(x)$ is the field in $\Omega$ describing drift, U is a task-related cost functional constructed by accumulating point costs (Fc(x)) along the path of the agent from the starting point to the target and Fc is a point cost function constructed in aim with the aspect of interest to the operator. In an energy exhaustive mission, it is desirable that the obstacle-free path connecting the starting and end points has a drift component that is in-phase with the direction along which motion is heading. A choice of Fc is (figure-4)

$$Fc = \frac{K}{2}(1-\cos(\Theta)) = \frac{K}{2}(1+\frac{\nabla V^T \Psi}{|\nabla V||\Psi|}). \quad (14)$$

A drift-sensitive descriptor is
$$P(x) = K - Fc = \frac{K}{2}(1-\frac{\nabla V^T \Psi}{|\nabla V||\Psi|}). \quad (15)$$

where $\Theta$ is the angle between $-\nabla V$ and $\Psi$ and K is a positive constant. Here P(x) is used to describe the utility of the drift at a certain point in space.

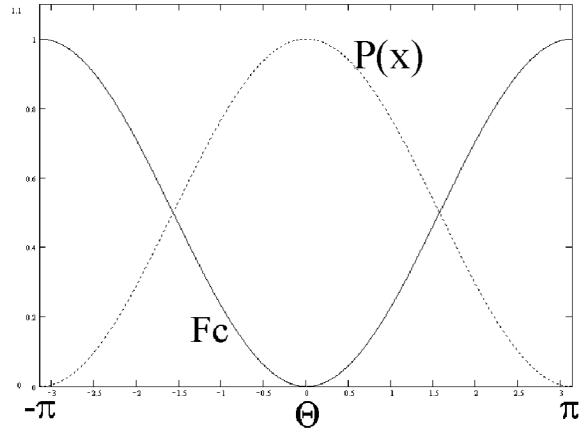

Figure-4: The suggested utility function versus $\Theta$.

The drift-sensitive GHPF operator is

$$\nabla^2 \mathbf{V} - \nabla \cdot (\frac{\nabla \mathbf{V}^T \Psi}{|\nabla \mathbf{V}||\Psi|})\nabla \mathbf{V}) \equiv \mathbf{0}. \quad (16)$$

and the overall drift-sensitive navigation policy generator is

$$\nabla^2 \mathbf{V} - \nabla \cdot (\frac{\nabla \mathbf{V}^T \Psi}{|\nabla \mathbf{V}||\Psi|})\nabla \mathbf{V}) \equiv \mathbf{0}. \quad x \in \Omega \quad (17)$$

subject to $V(x_S) = 1$, $V(x_T) = 0$, and $P(x) \equiv 0$ at $x \in O$, where O is the set of deterministic obstacles to be avoided. The path may be generated using the dynamical system in (7).

## V. Stability Analysis:

In this section, propositions along with their proofs are provided to explore the behavior of the GHPF approach suggested in section III.

**Proposition-1:** The path generated by the PDE-ODE system in (10) will avoid regions that have zero probability of achieving the task ($\Omega o$).

**Proof:** Assume a point x that is arbitrarily close to Ωo (figure-5). Since P(x) is differentiable, its value may be assumed equal to zero. Using the identity

$$\nabla \cdot (P(x)\nabla V(x)) = \nabla P(x)^t \nabla V(x) + P(x)\nabla^2 V(x) = 0 \tag{18}$$

when x is close to Ωo, equation 18 reduces to $- (-\nabla P(x)^t)(-\nabla V(x)) = 0$.

Note that $-\nabla P(x)$ points in the direction of increasing risk that leads to the region Ωo, while $-\nabla V(x)$ is the direction along which motion is to be steered. In other words, in the vicinity of Ωo, the planner will project motion tangentially to the boundary of the zero probability region. Hence, Ωo will be avoided.

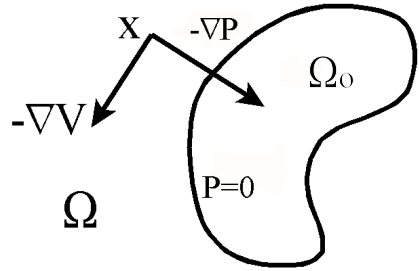

Figure-5: Zero-P regions will be avoided,

**Proposition-2:** A potential field that is generated using the boundary value problem in (10) cannot have any extrema local or global in its admissible workspace Ω-Ωo.

**Proof:** Note that from proposition-1, x will always stay in Ω-Ωo where P(x) is greater than zero. The fact that no minima or maxima can occur in Ω-Ωo may be inferred directly from the GHPF operator. Using the expansion in (12) and noting that at a critical point $\nabla V(x)=0$, we have

$$\nabla \cdot (P(x)\nabla V(x)) = P(x)\nabla^2 V(x) = P(x)\sum_{i=1}^{N} \frac{\partial^2 V(x)}{\partial x_i^2} = 0 \tag{19}$$

where $x=[x_1\ x_2\ ...\ x_N]^t$. At a local maxima, $\nabla V(x)$ will be negative in the whole local neighborhood surrounding x. This means that the second derivative of V(x) with respect to all $x_i$'s has to be positive (i.e. V(x) at the critical point is concaved upward). Since P(x) in that region is finite and positive, the governing relationship in (10) will be violated. The same thing will happen at a minima. $\nabla V(x)$ will be positive in the whole local neighborhood surrounding x. Therefore, no extrema of V local or global can occur in Ω-Ωo.

**Proposition-3:** For an $\Omega_a$ ($\Omega_a = \Omega - \Omega_o$) with a finite size, $V(x) \in \Omega_a$ is a Liapunov function candidate (LFC).

**Proof:** An LFC defined on a finite space must satisfy the following:

1- it must be at least continuous,

2- it must be positive in $\Omega_a$ ($V(x) > 0$, $x \in \Omega_a$),

3- its value must be zero at the target point ($V(x_T) = 0$).

Since $V(x)$ is forced to satisfy the differential condition in (10), it is analytic. Therefore, it satisfies the first condition. Since the global maximum happens at $x=x_s$ ($V(x_s)=1$) and a global minimum at $x=x_T$ ($V(x_T)=0$), the second and third conditions are satisfied.

**Proposition-4:** If $V(x)$ is constant at a subset of $\Omega$, it is constant for all $\Omega$.

**Proof:** First assume that $V(x) = C$ ( C is a constant) in $\omega$ where $\omega \subset \Omega$ (figure-6). Then consider an infinitesimally expanded region $\omega`$ that surrounds $\omega$. Finally, let $x_o$ be a point that lies on the boundary of $\omega$ ($\partial \omega$) and $x_o^+$ a point on $\partial \omega`$. Then, the potential at $x_o^+$ may be written as

$$V(x_o^+) = V(x_o) + dr \cdot (-\nabla V(x_o)^t n) \qquad (20)$$

where dr is a differential element and n is a unit vector normal to $\partial \omega$. Since V is constant inside $\omega$, the gradient field degenerates to zero. Since the continuity relation (10) is enforced in both $\omega$ and $\omega`$, equation (20) reduces to: $\qquad V(x_o^+) = V(x_o) = C$. $\qquad (21)$

By repeatedly applying the above procedure, it can be shown that the subregion $\omega$ may be expanded to include all $\Omega$. In other words, if $V(x)$ is constant on a subregion of $\Omega$, it will be constant for all $\Omega$.

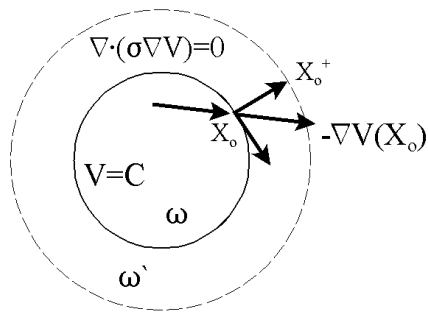

Figure-6: Subregions of degenerate fields cannot occur,

**Definition-1:** Let V(x) be at least twice differentiable scalar function (V(x): $R^N \rightarrow R$). A point $x_o$ is called a critical point of V if the gradient vanishes at that point ($\nabla V(x_o)=0$). Otherwise, $x_o$ is regular. A critical point is Morse if its Hessian matrix ($H(x_o)$) is nonsingular. V(x) is Morse if all of its critical points are Morse [34]. Showing that the GHPF is Morse is a key requirement for generating a provably-correct navigation policy [45]. The critical points in a Morse function are saddle points that do not trap motion and prevent the agent from reaching its target.

**Proposition-5:** If V(x) is a function defined in an N-dimensional space ($R^N$) on an open set $\Omega$ and satisfies (10), then the Hessian matrix at every critical point of V is nonsingular, i.e. V is Morse.

**Proof:** There are two properties of V that are used in the proof of proposition 5.

1- V(x) defined on an open set $\Omega$ contains no maxima or minima, local or global in $\Omega$. An extrema of V(x) can only occur at the boundary of $\Omega$,

2- if V(x) is constant in any open subset of $\Omega$, then it is constant for all $\Omega$.

Let $x_o$ be a critical point of V(x) inside $\Omega$. Since no maxima or minima of V exist inside $\Omega$, $x_o$ has to be a saddle point. Let V(x) be represented in the neighborhood of $x_o$ using a second order Taylor series:

$$V(x) = V(x_O) + \nabla V(x_O)^T (x - x_O) + \frac{1}{2}(x - x_O)^T H(x_O)(x - x_O) \qquad |x-x_o|<<1. \qquad (22)$$

Since Xo is a critical point of V, we have

$$V' = V(x) - V(x_O) = \frac{1}{2}(x - x_O)^T H(x_O)(x - x_O) \qquad |x-x_o|<<1. \qquad (23)$$

Notice that adding or subtracting a constant to V yields another potential field that satisfies the relation in (10). Using eigenvalue decomposition [35]

$$V' = \frac{1}{2}(x - x_O)^T U^T \begin{bmatrix} \lambda_1 & 0 & 0 & 0 \\ 0 & \lambda_2 & . & 0 \\ . & . & . & . \\ 0 & 0 & . & \lambda_N \end{bmatrix} U(x - x_O) = \frac{1}{2} \xi^T \begin{bmatrix} \lambda_1 & 0 & 0 & 0 \\ 0 & \lambda_2 & . & 0 \\ . & . & . & . \\ 0 & 0 & . & \lambda_N \end{bmatrix} \xi = \frac{1}{2}\sum_{i=1}^{N} \lambda_i \xi_i^2 \qquad (24)$$

where U is an orthonormal matrix of eigenvectors, $\lambda$'s are the eigenvalues of $H(x_o)$, and $\xi = [\xi_1 \, \xi_2 \, ..\xi_N]^T = U(x-x_o)$. V` cannot be zero on any open subset $\Omega$. Otherwise, it will be zero for all $\Omega$ which is not the case. This can only be true if and only if all the $\lambda_i$'s are nonzero. In other words, the Hessian of V at a critical point Xo is nonsingular. This makes V a Morse function.

**Proposition-6:** Let $V(x)$ be the potential field generated using the BVP in (10). The trajectory of the dynamical system

$$\dot{x} = -\nabla V(x) \tag{25}$$

will globally, asymptotically converge to: $\quad \lim_{t \to \infty} x \to x_T \quad\quad x(0) = x_s \in \Omega_a \tag{26}$

Proof of the above proposition is carried out using the LaSalle invariance principle [36].

**Proof:** Let $\Xi$ be the Liapunov function candidate

$$\Xi = V(x) \tag{27}$$

The time derivative (27) is $\quad \dot{\Xi}(x,\dot{x}) = K \cdot \nabla V(x)^T \dot{x} + \frac{1}{2} \dot{x}^T \dot{D}(x) \dot{x} + \dot{x}^T D(x) \ddot{x} . \tag{28}$

Substituting $\quad\quad\quad\quad\quad\quad\quad\quad \dot{x} = -\nabla V(x) \tag{29}$
in the above equation yields $\quad\quad\quad \dot{x} = -\|\nabla V(x)\|^2 \tag{30}$

$\nabla V$ will vanish at the target point $(x_T)$ and may have isolated critical points $\{x_i\}$ in $\Omega_a$. This results in

$$\dot{\Xi} \leq 0 \quad\quad \forall \; x . \tag{31}$$

According to the LaSalle principle, any bounded solution of (25) will converge to the minimum invariant set

$$E \subset \{x_T \cup \{x_i\}\} \; . \tag{32}$$

Determining E requires studying the critical points of $V(x)$ where $\nabla V(x)=0$. According to proposition-2, $x_T$ is the only minimum (stable equilibrium point) $V(x)$ can have. Besides $x_T$, $V(x)$ has other critical points $\{xc_i\}$ at which $\nabla V=0$. However, the Hessian at these points is non-singular, i.e. $V(x)$ is Morse. From the above, we conclude that E contains only one point $x = x_T$ to which motion will converge.

**Proposition-7:** Let $\rho(t)$ be the time parameterized trajectory generated by the gradient dynamical system

$$\dot{x} = -\nabla V(x) \quad\quad\quad x(0) = x_s \in \Omega_a \tag{33}$$

where $\rho(0) = x_s$, $\rho(\infty) = x_T$ and $V(x)$ is generated from the GHPF BVP in (10). Then $\rho(t)$ minimizes the cost

functional:

$$\int_0^\infty \frac{1}{P(x(t))} |\nabla V(x(t))| dt \qquad (34)$$

**Proof:** The proof follows directly from using electric current to model motion and using the conductivity field to model cost (cost=1/fitness). It is well-known that electric current will flow along the minimum resistance (cost) path. For a mathematical proof, that is based on the calculus of variation, see [43]. It is also worth mentioning that in addition to the above, the suggested navigation policy globally minimizes the energy functional

$$\int_\Omega P(x) |\nabla V(x)|^2 d\Omega \qquad (35)$$

which is known as the Dirichlet principle [44].

## VI. Results:

In this section, simulation results are presented to demonstrate the capabilities of the GHPF approach.

### V.1: From Probabilistic to Deterministic Planning:

There are several settings in which a harmonic potential field may be configured for navigation. Some of these settings are discussed in [31,32]. Each one of these configurations possesses distinct topological properties that are reflected in the integral and differential properties of the generated path. Figure-7 shows the control navigation policies of four different configurations for a simple rectangular environment.

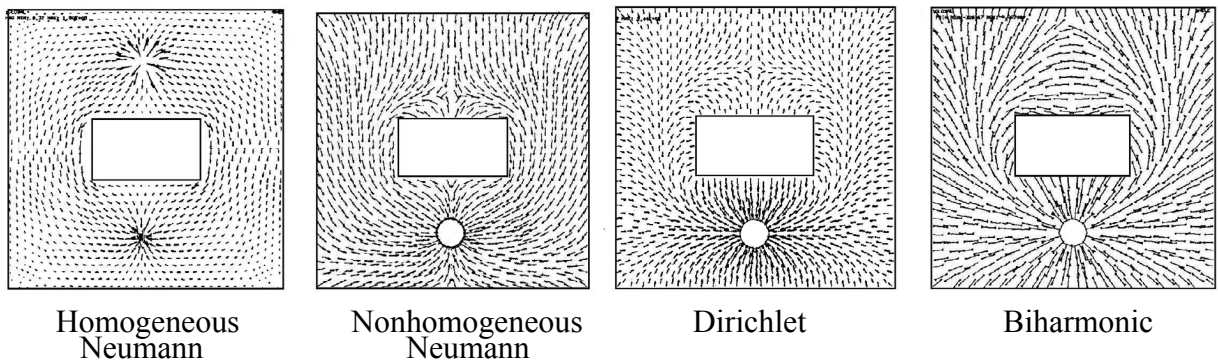

Homogeneous Neumann     Nonhomogeneous Neumann     Dirichlet     Biharmonic

Figure-7: Navigation policies, different settings, HPF approach.

The following example demonstrates the ability of the GHPF planner to handle deterministic environments by simply binary quantizing P(x) into 0 and 1(figure-8).

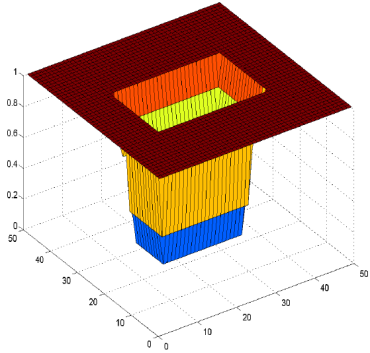
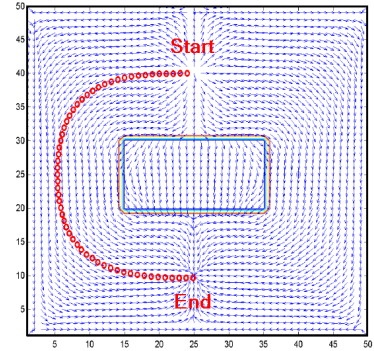

Figure-8: Probabilistic descriptor,     Figure-9: Path and navigation policy - GHPF approach,

The navigation control policy and the generated path are shown in figure-9. Unlike the deterministic case where the navigation policy degenerates inside the region to be avoided, the probabilistic HPF approach maintains the navigation field inside this region. This is of practical value because if a disturbance occurs throwing the robot inside a forbidden region, the robot can resume motion to the target instead of staying motionless in that area.

## V.2: Typical Non-divisible Environments:

In figure-10, a more challenging environment is used to test the approach. The probabilistic descriptor is shown as a 3-dimensional plot. As can be seen, segmenting this map into regions suitable for navigation, and into others that are not, is very difficult, if at all possible. Moreover, this binary segmentation could result in creating isolated regions that are disconnected from the rest of the workspace.

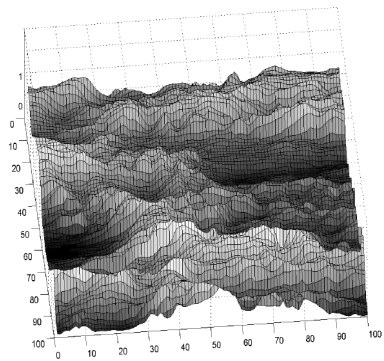

Figure-10: Probabilistic descriptor - surface map,

The navigation control policy is generated directly from the probabilistic map in figure-10 by the GHPF planner without any conditioning to the signal. The generated path superimposed on the image of the probabilistic descriptor (the brighter the area, the more suitable it is for navigation) of the environment is shown in figure-11. Visual assessment of the path in figure-11 clearly shows its smoothness and reasonable length. The path passes mainly through bright areas which are best suited for navigation. The corresponding navigation policy is shown in figure-12.

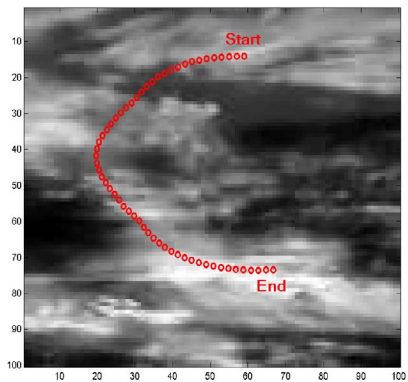 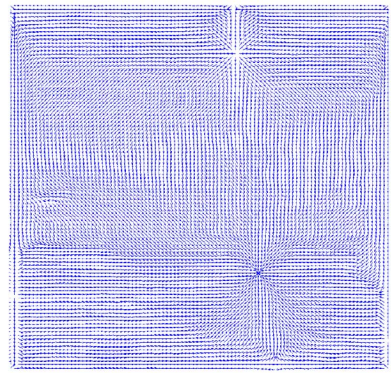

Figure-11: Generated path,   Figure-12: Navigation policy.

Figure-13 shows another nondivisible environment for different starting and end points.

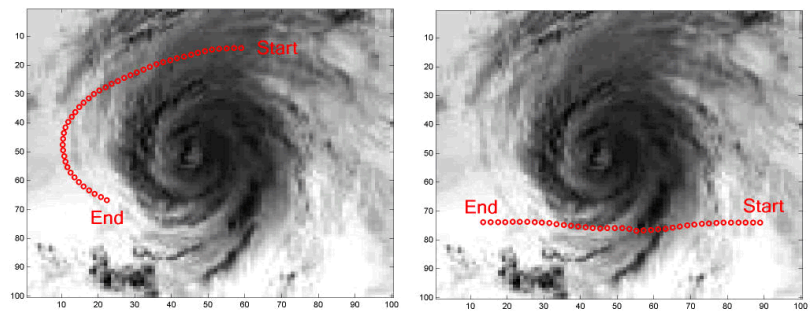

Figure-13: Different paths, same environment.

## V.3: Comparison With the D-Star Algorithm:

The GHPF approach is compared to the D-star algorithm [41,42]. The D-star is the benchmark for planning in the presence of uncertainties. Figure-14 shows an intensity map reflecting the fitness of space for motion. The brighter the region the more suitable it is for motion to pass through. The trajectory generated by the D-star is shown in figure-15. The trajectory generated by the GHPF planner along with an alternative path resulting from adding an obstruction (zero fitness zone) is shown in figure-16 . As can be seen, the trajectories from both techniques have comparable lengths and seek the most suitable area for motion to pass through. However, the path from the GHPF is considerably smoother than that produced by the D-star algorithm. As a result, the GHPF path is favorable for a dynamical agent to traverse. One must keep in mind that the GHPF approach is a goal-seeking planner where as the D-star is a path following planner. If a disturbance throws the agent off-course, an agent using GHPF will still move towards the target in an acceptable manner while an expensive re-planning has to be used in the case of the D-star.

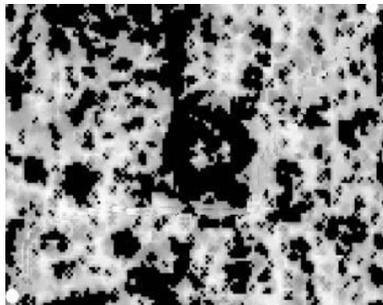 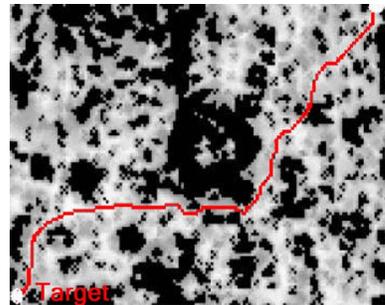

Figure-14: The probabilistic environment.    Figure-15: Path from the D-Star Algorithm

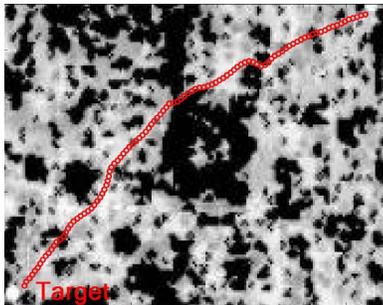 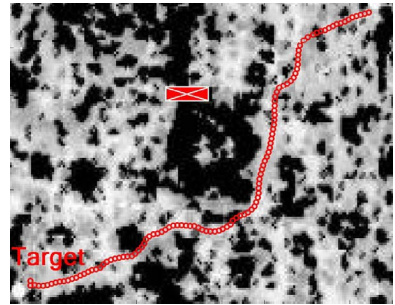

Figure-16: Paths from the GHPF planner.

### *V.4: Extreme Disorder:*

The capabilities of the GHPF planner in handling unstructured descriptors are tested using natural white noise as a representation of the environment (figure-17). Figure-18 shows the generated path superimposed on the image of the descriptor. One may observe the following:

1- the method works well although a highly unstructured, nondifferentiable white noise is used as the environment descriptor,

2- the path did converge to the target,

3- the path remained smooth with a reasonable length,

4- Although it is generally accepted that white noise means zero information, the path is low risk-sensitive. Figure-19 shows the path superimposed on two risk levels where the black areas of the first risk level are higher than 95% of the maximum of the environment and the other is higher than 85% of the maximum. As can be seen in both cases, the planner steered the path away from those regions.

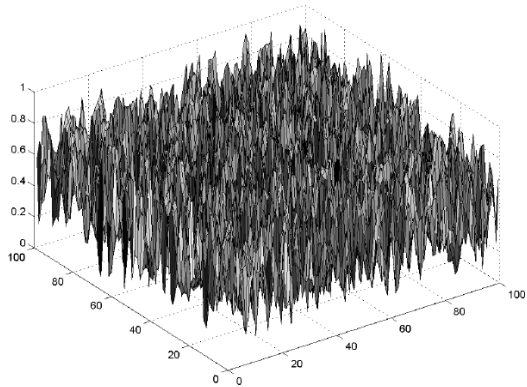
Figure-17: White noise,

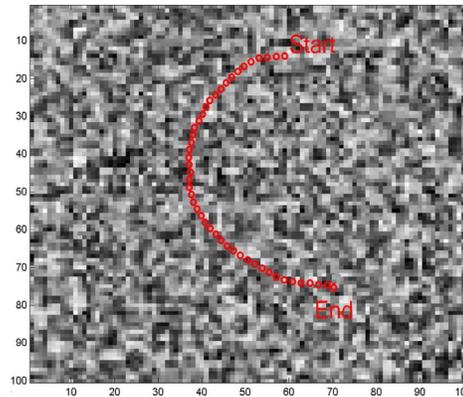
Figure-18: Path - random environment,

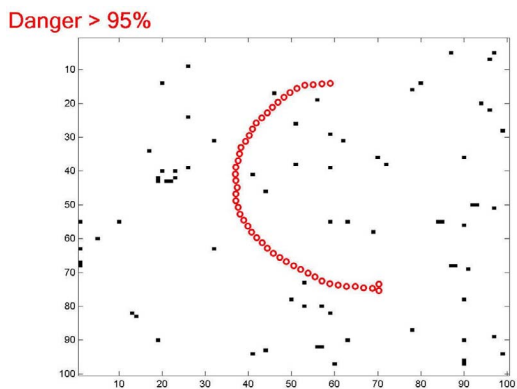

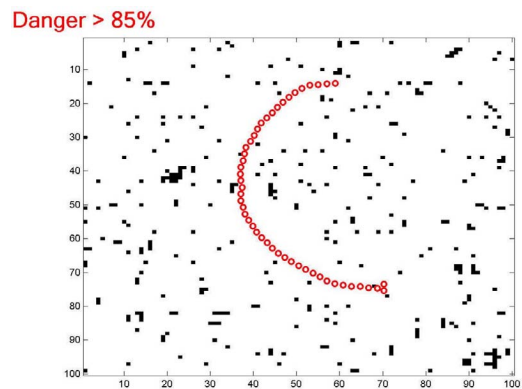

Figure-19: Path versus risk levels.

## *V.5: Drift-sensitive Planning*

The ability of the planner to process the drift data and the geometry of the space while generating a well-behaved navigation policy is demonstrated in different drift scenarios. In figure-20, the planner tackles a drift that has a vortex form and is rotating in a counter-clockwise direction. Two paths are shown; one with the box interior empty and one with an obstruction added. As can be seen in both cases, an obstacle-free trajectory that exploits the presence of drift in powering motion is laid to the target.

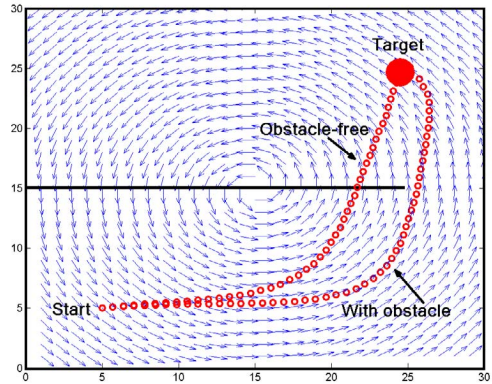

Figure-20: Trajectory in a ccw vortex

In figure-21, a random correlated drift field is used instead of the vortex field. As can be seen, a smooth path with reasonable length respecting the boundary of the space is laid to the target. Figure-22 shows the difference between the trajectory heading and the angle of drift at different points along the path. One can see that the majority of the drift components are aiding motion. Notice that motion starts from a difficult area where there is no direction to proceed along without going against drift.

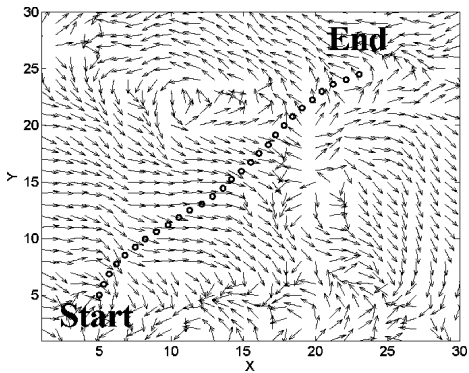 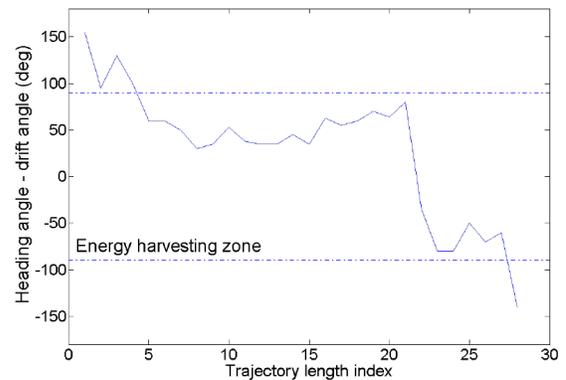

Figure-21: Trajectory in a random drift field,    Figure-22: Drift-heading differential in figure-21.

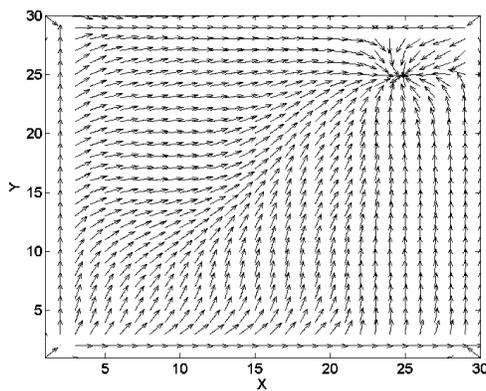

Figure-23: Navigation policy in figure-21.

Figure-23 shows the navigation policy corresponding to the drift in figure-21. Despite the random nature of drift, a smooth navigation policy is obtained for steering motion. In figure-24, an additional obstacle is added in the middle of the workspace for another case of random drift. As before, a smooth and safe path with reasonable length is generated. Figure-25 shows the difference between the heading of the trajectory and the angle of the drift. As can be seen, almost all the points of the trajectory lie in the energy harvesting zone with the absolute value of the difference being less than π/2.

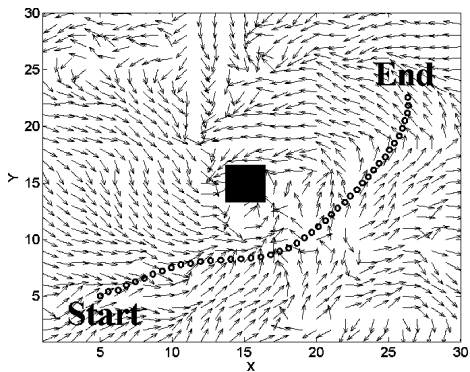
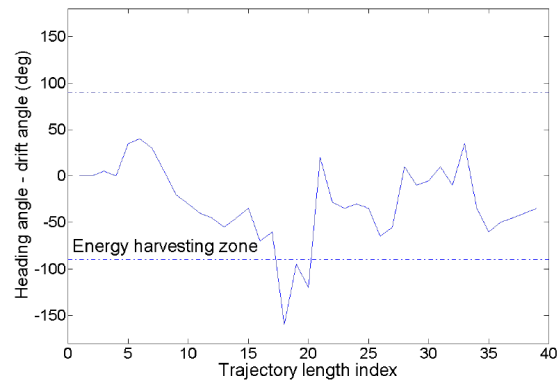

Figure-24: Trajectory in a random drift field.     Figure-25: Drift-heading differential in figure-24.

VII. Conclusion:

In this paper, the gamma-harmonic potential field approach is suggested for tackling planning in environments with inherent uncertainty that deny an operator the ability to segment the workspace into geometric regions of homogeneous attributes. An interpretation of the GHPF approach is also suggested for handling planning in cluttered environments that are populated by a drift field. The suggested extensions are proof of principle that the HPF approach is capable of efficiently addressing the information diversity issue needed for a planner to tackle a realistic situation in a provably-correct manner.

*Acknowledgment:* The author would like to thank King Fahad University of Petroleum and Minerals for its support of this work.